\documentclass[11pt]{article}
\usepackage{enumerate}
\usepackage{pdfsync}
\usepackage[OT1]{fontenc}

\usepackage[usenames]{color}
\usepackage{smile}
\usepackage[colorlinks,
            linkcolor=red,
            anchorcolor=blue,
            citecolor=blue
            ]{hyperref}
\usepackage{fullpage}
\usepackage[protrusion=true,expansion=true]{microtype}
\usepackage{pbox}
\usepackage{setspace}
\usepackage{tabularx}
\usepackage{float}
\usepackage{wrapfig,lipsum}
\usepackage{enumitem}
\usepackage{microtype}
\usepackage{graphicx}
\usepackage{subfigure}
\usepackage{pgfplots}
\usepackage{booktabs} 

\allowdisplaybreaks
\usepackage{colortbl}

\usepackage{xcolor}

\makeatletter
\newcommand*{\rom}[1]{\expandafter\@slowromancap\romannumeral #1@}
\makeatother
\title{\huge Solving PDEs With Deep Neural Nets under General Boundary Conditions}
\author
{
    Chenggong Zhang\thanks{Department of Electrical and Computer Engineering, University of California, Los Angeles, CA 90095, USA; email: {\tt chenggong61@g.ucla.edu}}~~~
}

\begin{document}
    \date{}
    \maketitle

\begin{abstract}
Partial Differential Equations (PDEs) are central to modeling complex systems across physical, biological, and engineering domains, yet traditional numerical methods often struggle with high-dimensional or complex problems. Physics-Informed Neural Networks (PINNs) have emerged as an efficient alternative by embedding physics-based constraints into deep learning frameworks, but they face challenges in achieving high accuracy and handling complex boundary conditions. In this work, we extend the Time-Evolving Natural Gradient (TENG) framework to address Dirichlet boundary conditions, integrating natural gradient optimization with numerical time-stepping schemes, including Euler and Heun methods, to ensure both stability and accuracy. By incorporating boundary condition penalty terms into the loss function, the proposed approach enables precise enforcement of Dirichlet constraints. Experiments on the heat equation demonstrate the superior accuracy of the Heun method due to its second-order corrections and the computational efficiency of Euler’s method for simpler scenarios. This work establishes a foundation for extending the framework to Neumann and mixed boundary conditions, as well as broader classes of PDEs, advancing the applicability of neural network-based solvers for real-world problems.
\end{abstract}

\section{Introduction}

Partial Differential Equations (PDEs) are foundational to modeling a wide range of physical, biological, and engineering systems. Solving PDEs is a cornerstone of computational science, but traditional numerical methods such as finite element and finite difference approaches can become computationally prohibitive, particularly in high-dimensional or complex domains. In recent years, Physics-Informed Neural Networks (PINNs)\citet{Raissi:PINN} have emerged as a promising alternative, leveraging the expressive power of deep learning to approximate solutions to PDEs by embedding the governing equations into the network's loss function.

Physics-Informed Neural Networks (PINNs) offer significant advantages, including being mesh-free and computationally efficient. These qualities make them particularly appealing for solving complex systems of equations like the Navier-Stokes equations, which have been a major focus of research in fluid dynamics. Despite significant progress over the past 50 years in the mathematical community, challenges remain: seamlessly incorporating noisy data into existing algorithms remains elusive, and generating meshes in high-dimensional spaces is inherently complex. In \citet{Cai:mesh_free}, the authors leveraged PINNs to address these challenges, demonstrating their effectiveness in solving a range of fluid dynamics problems. Remarkably, PINNs can also tackle ill-posed problems \citet{Cai:ill_posed}, which is typically infeasible using classical methods.

Despite their advantages of being mesh-free and computationally efficient, Physics-Informed Neural Networks (PINNs) have yet to achieve the level of accuracy offered by classical methods \citet{Grossmann:accuracy, Cuomo:accuracy}. This limitation is particularly evident in solving initial value problems in partial differential equations (PDEs), which are fundamental for modeling the evolution of dynamical systems. While machine learning techniques have shown promise in approximating solutions to complex partial differential equations (PDEs), they often struggle to achieve high accuracy, particularly when dealing with intricate initial conditions. This difficulty arises from the cumulative propagation of errors in PDE solvers over time, necessitating precise solutions at each time step to maintain overall accuracy.

Although various training strategies, such as global-in-time training \citet{Muller2023} and sequential-in-time training \citet{Chen2023a, Berman2023}, have been developed to mitigate these challenges, achieving consistent accuracy remains a significant obstacle to the practical application of PINNs. Addressing this, the authors in \citet{Chen:TENG} proposed the Time-Evolving Natural Gradient (TENG) method, which outperformed all previously benchmarked approaches, offering a promising advancement in the field.

In \citet{Chen:TENG}, the authors primarily focused on periodic boundary conditions. While effective in certain contexts, periodic boundary conditions are limited in scope. Non-periodic boundary conditions, such as Dirichlet or Neumann conditions, are essential for modeling a broader range of real-world phenomena where solutions are governed by fixed values or fluxes at the boundaries. These conditions enable accurate representation of complex systems, including heat transfer, fluid flow, and electromagnetic fields, where periodic assumptions are not applicable.

In this work, we extend the method proposed in \citet{Chen:TENG} to handle arbitrary boundary conditions, broadening its applicability to a larger class of PDE problems. This enhancement significantly increases its potential for addressing real-world challenges. To validate the effectiveness of our approach, we present a series of numerical experiments on benchmark PDE problems of varying complexity.

\section{Related Work}

\subsection{Machine Learning for Solving PDEs}
Machine learning has emerged as a promising tool for solving partial differential equations (PDEs), utilizing neural networks as function approximators to model solutions. Broadly, there are two categories of approaches: global-in-time optimization and sequential-in-time optimization.

Global-in-time optimization, such as the Physics-Informed Neural Network (PINN) framework \citet{Raissi2019}, aims to optimize a neural network representation of the solution over the entire time and spatial domain simultaneously. Variants of this approach include the Deep Ritz Method \citet{Weinan2017}, which leverages the variational form of the PDE when it exists. While effective in many cases, these methods can struggle with maintaining accuracy, particularly in problems with intricate boundary conditions, where incorporating boundary constraints seamlessly remains challenging.

In contrast, sequential-in-time optimization, also known as the neural Galerkin method, focuses on updating the neural network representation step-by-step in time. Methods in this category include the Time-Dependent Variational Principle (TDVP) \citet{Dirac1930, Koch2007, Carleo2017, Du2021, Berman2023} and Optimization-Based Time Integration (OBTI) \citet{Kochkov2018, Gutierrez2022}. Sequential-in-time strategies are particularly advantageous in handling evolving boundary conditions, as they allow for explicit incorporation of such conditions at each time step.

Beyond these direct PDE-solving strategies, data-driven approaches have also been employed to model PDEs based on observed data. These methods include Neural ODEs \citet{Chen2018}, Graph Neural Networks \citet{Pfaff2020, SanchezGonzalez2020}, Neural Fourier Operators \citet{Li2020}, and DeepONet \citet{Lu2019}. While these techniques have demonstrated strong performance in approximating PDE solutions, adapting them to accurately and efficiently enforce complex boundary conditions remains a significant challenge.

\subsection{Natural Gradient Optimization}
Natural gradient optimization, introduced by Amari \citet{Amari1998}, has become a cornerstone in second-order optimization techniques for machine learning. By adjusting the gradient direction based on the Fisher information matrix, natural gradient descent incorporates the geometry of the parameter space, leading to faster and more stable convergence compared to standard gradient methods. This feature makes it particularly effective for problems with complex constraints, such as boundary conditions in PDEs.

Natural gradient methods have been widely applied in neural network optimization \citet{Pascanu2013}, reinforcement learning \citet{Kakade2001}, quantum optimization, and PINN training \citet{Muller2023}. In the context of solving PDEs, these methods can improve the enforcement of boundary conditions by effectively navigating the optimization landscape shaped by the PDE constraints and boundary terms. Recent works have demonstrated the utility of natural gradients in PINN training \citet{Muller2023}, highlighting their potential to address issues such as cumulative error propagation and the incorporation of boundary constraints.

\subsection{Boundary Conditions in PDEs and Neural Networks}
Handling boundary conditions is a critical challenge in applying machine learning to PDEs. Global-in-time methods, such as PINNs, typically encode boundary conditions directly into the loss function. However, for complex or non-periodic boundary conditions, achieving accurate enforcement can become computationally challenging. Sequential-in-time methods, such as TDVP and OBTI, provide a more flexible framework for integrating boundary conditions, as they allow conditions to be explicitly updated at each time step. This stepwise approach is particularly suitable for problems involving dynamic or mixed boundary conditions, where the constraints may evolve over time.

Recent advancements in natural gradient optimization offer an additional layer of robustness in solving PDEs with boundary constraints. By leveraging second-order information, natural gradient methods can better balance the interplay between minimizing PDE residuals and satisfying boundary conditions, making them well-suited for addressing problems with intricate or non-periodic boundaries.

\section{Methodology}

The Time-Evolving Natural Gradient (TENG) framework \citet{Chen:TENG} combines the time-dependent variational principle with optimization-based time integration using the natural gradient. This approach leverages second-order optimization to enhance convergence and numerical stability. While TENG has demonstrated effectiveness under periodic boundary conditions, it does not generalize to problems with arbitrary boundary conditions, such as Dirichlet or Neumann conditions. In this work, we extend the TENG framework to address these limitations.

\subsection{Generalized Loss Functions}

To handle Dirichlet and Neumann boundary conditions, we modify the TENG framework by incorporating penalty terms into the loss function. These terms enforce the boundary conditions in addition to minimizing the PDE residuals.

\subsubsection{Dirichlet Boundary Conditions}
For Dirichlet boundary conditions, where the solution is constrained to fixed values at the domain boundaries, the total loss function is defined as:
\[
\mathcal{L}_{\text{Dirichlet}} = \mathcal{L}_{\text{PDE}} + \lambda_{\text{Dirichlet}} \| u(x_{\text{boundary}}) - u_{\text{Dirichlet}} \|^2,
\]
where:
\begin{itemize}
    \item \( \mathcal{L}_{\text{PDE}} \): The residual loss of the governing PDE, ensuring compliance with the physical laws.
    \item \( \| u(x_{\text{boundary}}) - u_{\text{Dirichlet}} \|^2 \): A penalty term enforcing the prescribed Dirichlet boundary values.
    \item \( \lambda_{\text{Dirichlet}} \): A weighting factor controlling the relative importance of the boundary condition term.
\end{itemize}

This formulation ensures that the solution satisfies both the PDE and the Dirichlet boundary conditions.

\subsection{Optimization Process}

The optimization process iteratively minimizes the total loss function for each time step of the PDE solution. This is achieved using the natural gradient, which projects the gradient of the loss function in the \(u\)-space onto the parameter \(\theta\)-space. By formulating the parameter updates as a least-squares problem, the natural gradient method achieves second-order optimization, improving both stability and convergence rates.

The details of the parameter update process are implemented in the \texttt{TENG\_stepper} subroutine, described in \ref{alg:teng_stepper}.

\begin{algorithm}[tb]
   \caption{\texttt{TENG\_stepper}: Parameter Update Subroutine}
   \label{alg:teng_stepper}
\begin{algorithmic}
   \STATE {\bfseries Input:} Initial parameters \(\theta_{\text{init}}\), target solution \(u_{\text{target}}\)
   \STATE Initialize \(n \gets 0\), \(\theta \gets \theta_{\text{init}}\)
   \REPEAT
       \STATE Compute the gradient: \(\Delta u(x) \gets -\alpha_n \frac{\partial L(\hat{u}_\theta, u_{\text{target}})}{\partial \hat{u}_\theta}\)
       \STATE Compute the Jacobian: \(J(x)_j \gets \frac{\partial \hat{u}_\theta(x)}{\partial \theta_j}\)
       \STATE Solve the least-squares problem:
       \[
       \Delta \theta = \arg \min_{\Delta \theta \in \mathbb{R}^{N_p}} \left\| \Delta u(\cdot) - \sum_j J(\cdot)_j \Delta \theta_j \right\|^2_{L^2(X)}
       \]
       \STATE Update parameters: \(\theta \gets \theta + \Delta \theta\)
       \STATE Increment iteration: \(n \gets n + 1\)
   \UNTIL{$n \geq N_{\text{it}}$}
   \STATE {\bfseries Output:} \(\theta\)
\end{algorithmic}
\end{algorithm}

\subsection{Time-Stepping Schemes}

The \texttt{TENG\_stepper} subroutine is integrated into two numerical time-stepping schemes: Euler and Heun methods. These schemes iteratively evolve the parameters \(\theta\) at each time step while ensuring numerical stability and accuracy.

\subsubsection{Euler Method}
Euler's method is a first-order integration scheme that updates the solution at each time step based on the PDE residual. The algorithm is summarized in \ref{alg:teng_euler}.

\begin{algorithm}[tb]
   \caption{TENG\_Euler: A 1st-Order Integration Scheme}
   \label{alg:teng_euler}
\begin{algorithmic}
   \STATE {\bfseries Input:} Initial parameters \(\theta_{t=0}\), time step \(\Delta t\), total time \(T\)
   \STATE Initialize \(t \gets 0\)
   \REPEAT
       \STATE Compute \(u_{\text{target}}(x) \gets \hat{u}_{\theta_t}(x) + \Delta t L \hat{u}_{\theta_t}(x)\)
       \STATE Update \(\theta_{t+\Delta t} \gets \text{TENG\_stepper}(\theta_t, u_{\text{target}})\)
       \STATE Increment \(t \gets t + \Delta t\)
   \UNTIL{$t \geq T$}
   \STATE {\bfseries Output:} Final parameters \(\theta_{t=T}\)
\end{algorithmic}
\end{algorithm}

\subsubsection{Heun Method}
Heun's method is a second-order integration scheme that incorporates an intermediate step to improve accuracy. The algorithm is summarized in \ref{alg:teng_heun}.

\begin{algorithm}[tb]
   \caption{TENG\_Heun: A 2nd-Order Integration Scheme}
   \label{alg:teng_heun}
\begin{algorithmic}
   \STATE {\bfseries Input:} Initial parameters \(\theta_{t=0}\), time step \(\Delta t\), total time \(T\)
   \STATE Initialize \(t \gets 0\)
   \REPEAT
       \STATE Compute \(u_{\text{temp}}(x) \gets \hat{u}_{\theta_t}(x) + \Delta t L \hat{u}_{\theta_t}(x)\)
       \STATE Update \(\theta_{\text{temp}} \gets \text{TENG\_stepper}(\theta_t, u_{\text{temp}})\)
       \STATE Compute \(u_{\text{target}}(x) \gets \hat{u}_{\theta_t}(x) + \frac{\Delta t}{2} \big(L \hat{u}_{\theta_t}(x) + L \hat{u}_{\theta_{\text{temp}}}(x)\big)\)
       \STATE Update \(\theta_{t+\Delta t} \gets \text{TENG\_stepper}(\theta_{\text{temp}}, u_{\text{target}})\)
       \STATE Increment \(t \gets t + \Delta t\)
   \UNTIL{$t \geq T$}
   \STATE {\bfseries Output:} Final parameters \(\theta_{t=T}\)
\end{algorithmic}
\end{algorithm}

\subsection{Numerical Experiments}

We validated the proposed framework by solving the heat equation with Dirichlet boundary conditions. The experiments compared the Euler and Heun methods in terms of accuracy, stability, and convergence. The setup included:
\begin{enumerate}
    \item Solving the heat equation on a circular domain with analytical solutions derived from Bessel functions.
    \item Testing various initial conditions to evaluate robustness and precision.
\end{enumerate}

The results demonstrated that the Heun method achieves higher accuracy due to its second-order corrections, while the Euler method provides a computationally efficient alternative for simpler scenarios.

\section{Experiments}

We tested our algorithm on the heat equation with Dirichlet boundary conditions. The heat equation is a fundamental partial differential equation that models the distribution of heat (or diffusion of substances) in a given region over time, making it essential in fields like physics, engineering, and materials science. Its mathematical framework also provides key insights into broader areas such as probability theory, financial modeling, and the study of partial differential equations in general. Specifically, let $\Omega = B(0, 1)\subset \mathbb{R}^2$ be the open disk centered on $0$ with radius 1. We consider the two-dimensional isotropic heat equation $$\frac{\partial u}{\partial t} = \nu (\frac{\partial^2 u}{\partial x_1^2} + \frac{\partial^2 u}{\partial x_2^2}) $$ with a diffusivity constant $\nu = 1/10$. We also impose the Dirichlet boundary condition that $$u(x, t) = 0, \forall x\in \partial \Omega.$$ Since the heat equation on a disk has an analytical solution with Bessel functions as the basis, we consider initial conditions as linear combinations of Bessel functions. This approach provides precise analytical solutions, making it ideal for benchmarking.

\subsection{Experiment 1}
We consider the initial condition

\begin{align*}
   u_0(r, \theta) &= \frac{1}{4}(Z_{01}(r, \theta) - \frac{1}{4}Z_{02}(r, \theta) + \frac{1}{16}Z_{03}(r, \theta)\\
     &- \frac{1}{64}Z_{04}(r, \theta) + Z_{11}(r, \theta) - \frac{1}{2}Z_{12}(r, \theta) + \frac{1}{4}Z_{13}(r, \theta)\\
     &- \frac{1}{8}Z_{14}(r, \theta) + Z_{21}(r, \theta) + Z_{31}(r, \theta) + Z_{41}(r, \theta))  
\end{align*}

where $r$ and $\theta$ are the polar coordinate variables and $Z_{mn}$ represent the disk harmonics defined as $Z_{mn}(r, \theta) = J_{m}(\lambda_{nm} r)\cos(m\theta)$ with $J_m$ the $m$th Bessel function and $\lambda_{mn}$ the $n$th zero of the $m$th Bessel function.

\begin{figure}[h]
    \centering
    \includegraphics[width=0.5\textwidth]{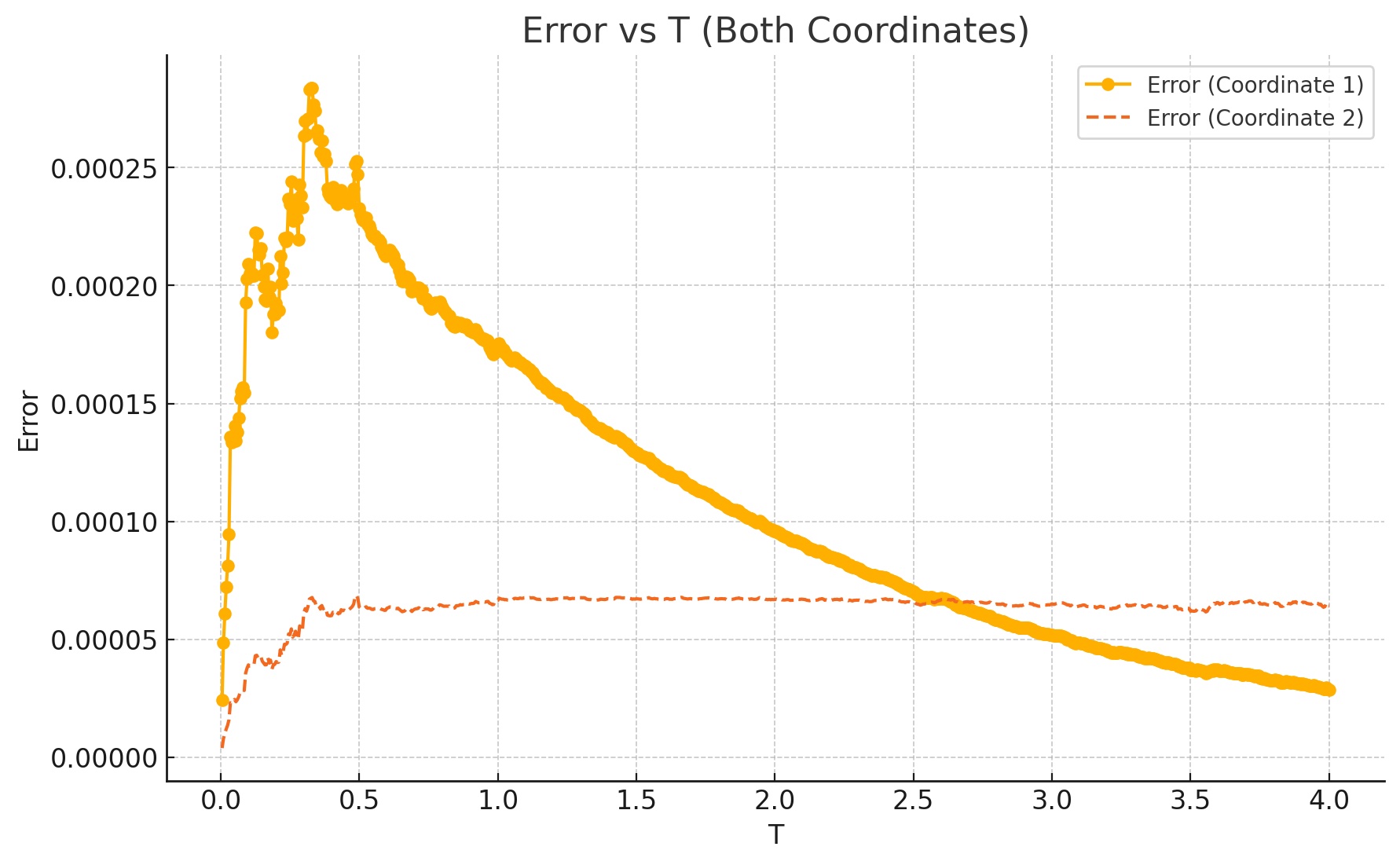}
    \caption{TENG\_Heun Result}
    \label{fig:init_huen}
\end{figure}

\begin{figure}[h]
    \centering
    \includegraphics[width=0.5\textwidth]{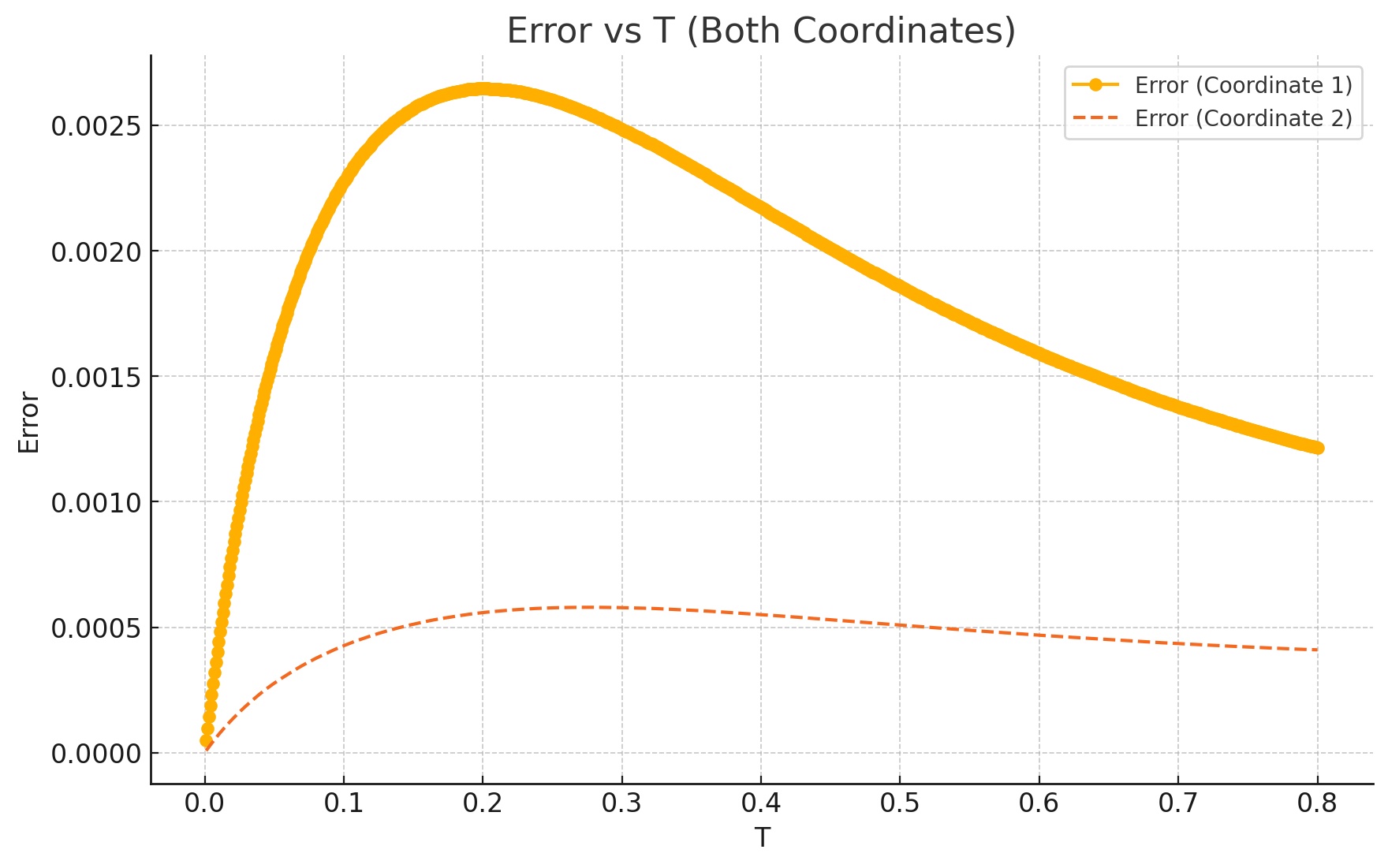}
    \caption{TENG\_Euler Result}
    \label{fig:init_euler}
\end{figure}

We solve this equation on the domain $B(0, 1) \times [0, \mathcal{T}]$, where $\mathcal{T} = 4$, using TENG\_Heun methods with a stepsize of $h = 0.005$. We also output the result on the same domain with $\mathcal{T} = 0.8$ using TENG\_Euler with the same stepsize $h = 0.005$.
The resulting errors of TENG\_Heun and TENG\_Euler are visualized in Figure \ref{fig:init_huen} and Figure \ref{fig:init_euler} respectively.

To initialize the neural network, we utilized a set of supervised pre-trained weights. Our results show that the error for TENG Euler is significantly larger compared to TENG Heun. Notably, TENG Heun demonstrates promising accuracy, with error already within $1e-4$. We expect more improvement is possible if using a higher-order integration scheme, such as RK4.

\subsection{Experiment 2}

We consider the same initial condition

\begin{align*}
    u_0(r, \theta)
    &= \frac{1}{4}(Z_{01}(r, \theta) - \frac{1}{4}Z_{02}(r, \theta) + \frac{1}{16}Z_{03}(r, \theta) - \frac{1}{64}Z_{04}(r, \theta) \\
     &+ Z_{11}(r, \theta) - \frac{1}{2}Z_{12}(r, \theta) + \frac{1}{4}Z_{13}(r, \theta)\\
     &- \frac{1}{8}Z_{14}(r, \theta) + Z_{21}(r, \theta) + Z_{31}(r, \theta) + Z_{41}(r, \theta))
\end{align*}

Once again, we solve this equation on the domain $B(0, 1) \times [0, \mathcal{T}]$, where $\mathcal{T} = 4$ using the TENG\_Huen method and $\mathcal{T} = 0.8$ using the TENG\_Euler method. However, instead of importing a set of pre-trained weights for neural network initialization, we designed the model to consist of two neural networks. The model's output is computed as the difference between the two neural networks, plus the initial condition.

During initialization, both neural networks were set to have equal weights. During training, the parameters of one neural network were frozen, and only the parameters of the other network were updated. This approach resulted in significantly larger errors compared to the pre-trained initialization.

These results highlight the critical importance of the initialization phase in guiding the model towards the correct solution. The choice of initial parameters plays a pivotal role in ensuring effective training and reducing error.

\section{Discussion}
In this work, we generalized periodic boundary conditions in \citet{Chen:TENG} to Dirichlet boundary conditions, demonstrating the adaptability of our approach to handle different types of constraints. This extension broadens the applicability of the method to a wider range of physical and mathematical problems, where Dirichlet boundary conditions play a critical role. The results underscore the potential of this generalized framework for accurately solving boundary value problems while maintaining computational efficiency and flexibility.

Looking ahead, several exciting directions emerge for future research. An immediate extension would be the incorporation of Neumann boundary conditions, which model flux-constrained scenarios and are highly relevant in heat transfer, fluid dynamics, and electromagnetics. Exploring mixed boundary conditions, where different segments of the boundary enforce Dirichlet or Neumann constraints, would further enhance the utility of the approach for real-world applications.

Additionally, extending the current schema to a broader class of partial differential equations (PDEs) beyond initial-value problems would be an important advancement. Addressing PDEs that involve complex geometries, nonlinearity, or time-dependent boundaries could significantly expand the scope of the method. Such generalizations would not only enhance the mathematical framework but also provide critical tools for scientific and engineering applications, ranging from materials science to climate modeling.

These advancements would represent significant contributions to the field, offering both theoretical insights and practical tools for solving challenging problems across disciplines.

\section{Acknowledgments}

I would like to thank Professor Di Luo and Zhuo Chen, their valuable suggestions and great support throughout this project. This note was not reviewed, approved, or endorsed by Prof. Di Luo or Zhuo Chen. Any errors are solely the author's responsibility. I would like to thank Xinjie He for her valuable contribution.
\clearpage
\appendix
\onecolumn
\section{Hyperparamter}
\begin{table}[htbp]
\centering
\caption{Hyperparameters for Euler and Heun Methods}
\label{tab:hyperparameters}
\begin{tabular}{|l|c|c|}
\hline
\textbf{Hyperparameter}         & \textbf{Euler Method} & \textbf{Heun Method} \\ \hline
\texttt{D}                      & 0.1                  & 0.1                  \\ \hline
\texttt{Equation}               & Heat                 & Heat                 \\ \hline
\texttt{Number of Steps}        & 800                  & 800                  \\ \hline
\texttt{Iterations per Step}    & 5                    & 5                    \\ \hline
\texttt{Time Step (\texttt{dt})} & 0.001                & 0.005                \\ \hline
\texttt{Integrator}             & Euler                & Heun                 \\ \hline
\texttt{Save Directory}         & None                 & None                 \\ \hline
\texttt{Model State}            & \texttt{model\_state\_circ\_2d.pickle} & \texttt{model\_state\_circ\_2d.pickle} \\ \hline
\texttt{Model Seed}             & 1234                 & 1234                 \\ \hline
\texttt{Number of Samples}      & 65536                & 65536                \\ \hline
\texttt{Sampler Seed}           & 4321                 & 4321                 \\ \hline
\texttt{Policy Gradient Params} & 1536                 & 1536                 \\ \hline
\texttt{Policy Gradient Seed}   & 8844                 & 8844                 \\ \hline
\texttt{Policy Gradient 2 Params} & 1024               & 1024                 \\ \hline
\texttt{Policy Gradient 2 Seed} & 8848                 & 8848                 \\ \hline
\end{tabular}
\end{table}

\clearpage
\bibliography{arxiv}

@article{Raissi:PINN,
  author       = {Maziar Raissi and Paris Perdikaris and George E. Karniadakis},
  title        = {Physics Informed Deep Learning (Part {I):} Data-driven Solutions of Nonlinear Partial Differential Equations},
  journal      = {CoRR},
  volume       = {abs/1711.10561},
  year         = {2017},
  url          = {https://arxiv.org/abs/1711.10561},
  eprinttype   = {arXiv},
  eprint       = {1711.10561},
}

@misc{Chen:TENG,
  title         = {TENG: Time-Evolving Natural Gradient for Solving PDEs With Deep Neural Nets Toward Machine Precision},
  author        = {Zhuo Chen and Jacob McCarran and Esteban Vizcaino and Marin Solja{\v c}i{\'c} and Di Luo},
  year          = {2024},
  eprint        = {2404.10771},
  archivePrefix = {arXiv},
  primaryClass  = {cs.LG},
  url           = {https://arxiv.org/abs/2404.10771},
}

@misc{Grossmann:accuracy,
  title         = {Can Physics-Informed Neural Networks beat the Finite Element Method?},
  author        = {Tamara G. Grossmann and Urszula Julia Komorowska and Jonas Latz and Carola-Bibiane Sch{\"o}nlieb},
  year          = {2023},
  eprint        = {2302.04107},
  archivePrefix = {arXiv},
  primaryClass  = {math.NA},
  url           = {https://arxiv.org/abs/2302.04107},
}

@article{Cai:ill_posed,
  author  = {Cai, Shengze and Wang, Zhicheng and Wang, Sifan and Perdikaris, Paris and Karniadakis, George Em},
  title   = {Physics-Informed Neural Networks for Heat Transfer Problems},
  journal = {Journal of Heat Transfer},
  volume  = {143},
  number  = {6},
  pages   = {060801},
  year    = {2021},
  doi     = {10.1115/1.4050542},
  url     = {https://doi.org/10.1115/1.4050542},
}

@article{Cuomo:accuracy,
  title     = {Scientific machine learning through physics--informed neural networks: Where we are and what’s next},
  author    = {Cuomo, Salvatore and Di Cola, Vincenzo Schiano and Giampaolo, Fabio and Rozza, Gianluigi and Raissi, Maziar and Piccialli, Francesco},
  journal   = {Journal of Scientific Computing},
  volume    = {92},
  number    = {3},
  pages     = {88},
  year      = {2022},
  publisher = {Springer},
}

@misc{Cai:mesh_free,
  title         = {Physics-informed neural networks (PINNs) for fluid mechanics: A review},
  author        = {Shengze Cai and Zhiping Mao and Zhicheng Wang and Minglang Yin and George Em Karniadakis},
  year          = {2021},
  eprint        = {2105.09506},
  archivePrefix = {arXiv},
  primaryClass  = {physics.flu-dyn},
  url           = {https://arxiv.org/abs/2105.09506},
}

@misc{Muller2023,
  title         = {Achieving High Accuracy with PINNs via Energy Natural Gradients},
  author        = {Johannes M{\"u}ller and Marius Zeinhofer},
  year          = {2023},
  eprint        = {2302.13163},
  archivePrefix = {arXiv},
  primaryClass  = {cs.LG},
  url           = {https://arxiv.org/abs/2302.13163},
}

@inproceedings{Chen2023a,
  title     = {Implicit Neural Spatial Representations for Time-dependent PDEs},
  author    = {Chen, Honglin and Wu, Rundi and Grinspun, Eitan and Zheng, Changxi and Chen, Peter Yichen},
  booktitle = {International Conference on Machine Learning},
  pages     = {5162--5177},
  year      = {2023},
  organization = {PMLR},
}

@misc{Berman2023,
  title         = {Randomized Sparse Neural Galerkin Schemes for Solving Evolution Equations with Deep Networks},
  author        = {Jules Berman and Benjamin Peherstorfer},
  year          = {2023},
  eprint        = {2310.04867},
  archivePrefix = {arXiv},
  primaryClass  = {cs.LG},
  url           = {https://arxiv.org/abs/2310.04867},
}

@article{Amari1998,
  title   = {Natural gradient works efficiently in learning},
  author  = {Amari, Shun-ichi},
  journal = {Neural Computation},
  volume  = {10},
  number  = {2},
  pages   = {251--276},
  year    = {1998},
}

@article{Carleo2017,
  title     = {Solving the quantum many-body problem with artificial neural networks},
  author    = {Carleo, Giuseppe and Troyer, Matthias},
  journal   = {Science},
  volume    = {355},
  number    = {6325},
  pages     = {602--606},
  year      = {2017},
  publisher = {American Association for the Advancement of Science},
}

@inproceedings{Chen2018,
  title     = {Neural ordinary differential equations},
  author    = {Chen, Ricky T. Q. and Rubanova, Yulia and Bettencourt, Jesse and Duvenaud, David K.},
  booktitle = {Advances in Neural Information Processing Systems},
  volume    = {31},
  year      = {2018},
}

@article{Dirac1930,
  title   = {Note on exchange phenomena in the Thomas atom},
  author  = {Dirac, Paul Adrien Maurice},
  journal = {Mathematical Proceedings of the Cambridge Philosophical Society},
  volume  = {26},
  pages   = {376--385},
  year    = {1930},
  publisher = {Cambridge University Press},
}

@article{Du2021,
  title   = {Evolutional deep neural network},
  author  = {Du, Yaowei and Zaki, Tamer A.},
  journal = {Physical Review E},
  volume  = {104},
  number  = {4},
  year    = {2021},
  doi     = {10.1103/physreve.104.045303},
}

@article{Gutierrez2022,
  title   = {Real time evolution with neural-network quantum states},
  author  = {Gutierrez, I. L. and Mendl, C. B.},
  journal = {Quantum},
  volume  = {6},
  pages   = {627},
  year    = {2022},
}

@inproceedings{Kakade2001,
  title     = {A natural policy gradient},
  author    = {Kakade, Sham M.},
  booktitle = {Advances in Neural Information Processing Systems},
  volume    = {14},
  year      = {2001},
}

@article{Koch2007,
  title   = {Dynamical low-rank approximation},
  author  = {Koch, Oliver and Lubich, Christian},
  journal = {SIAM Journal on Matrix Analysis and Applications},
  volume  = {29},
  number  = {2},
  pages   = {434--454},
  year    = {2007},
}

@article{Kochkov2018,
  title   = {Variational optimization in the AI era: Computational graph states and supervised wavefunction optimization},
  author  = {Kochkov, Dmitry and Clark, Bryan K.},
  journal = {arXiv preprint arXiv:1811.12423},
  year    = {2018},
}

@article{Li2020,
  title   = {Fourier neural operator for parametric partial differential equations},
  author  = {Li, Zongyi and Kovachki, Nikola and Azizzadenesheli, Kamyar and Liu, Burigede and Bhattacharya, Kaushik and Stuart, Andrew and Anandkumar, Anima},
  journal = {arXiv preprint arXiv:2010.08895},
  year    = {2020},
}

@misc{Lu2019,
  title         = {DeepONet: Learning nonlinear operators for identifying differential equations based on the universal approximation theorem of operators},
  author        = {Lu, Lu and Jin, Pengzhan and Karniadakis, George Em},
  year          = {2019},
  eprint        = {1910.03193},
  archivePrefix = {arXiv},
  url           = {https://arxiv.org/abs/1910.03193},
}

@inproceedings{Raissi2019,
  title   = {Physics-informed neural networks: A deep learning framework for solving forward and inverse problems involving nonlinear partial differential equations},
  author  = {Raissi, Maziar and Perdikaris, Paris and Karniadakis, George E.},
  booktitle = {Journal of Computational Physics},
  volume  = {378},
  pages   = {686--707},
  year    = {2019},
}

@misc{Weinan2017,
      title={The Deep Ritz method: A deep learning-based numerical algorithm for solving variational problems}, 
      author={Weinan E and Bing Yu},
      year={2017},
      eprint={1710.00211},
      archivePrefix={arXiv},
      primaryClass={cs.LG},
      url={https://arxiv.org/abs/1710.00211}, 
}

@inproceedings{Pfaff2020,
  title={Learning mesh-based simulation with graph networks},
  author={Pfaff, Tobias and Fortunato, Meire and Sanchez-Gonzalez, Alvaro and Battaglia, Peter},
  booktitle={International conference on learning representations},
  year={2020}
}

@inproceedings{Pascanu2013,
  title={On the difficulty of training recurrent neural networks},
  author={Pascanu, Razvan and Mikolov, Tomas and Bengio, Yoshua},
  booktitle={International conference on machine learning},
  pages={1310--1318},
  year={2013},
  organization={Pmlr}
}

@inproceedings{SanchezGonzalez2020,
  title={Learning to simulate complex physics with graph networks},
  author={Sanchez-Gonzalez, Alvaro and Godwin, Jonathan and Pfaff, Tobias and Ying, Rex and Leskovec, Jure and Battaglia, Peter},
  booktitle={International conference on machine learning},
  pages={8459--8468},
  year={2020},
  organization={PMLR}
}
\bibliographystyle{ims}

\end{document}